# Attention-based Assisted Excitation for Salient Object Detection

Saeed Masoudnia [a], Melika Kheirieh [a], Abdol-Hossein Vahabie [b], Babak Nadjar Araabi [a]

[a] Machine Learning and Computational Modelling Lab, Control and Intelligent Processing Center of Excellence, School of Electrical and Computer Engineering, College of Engineering, University of Tehran, Tehran, Iran.
[b] School of Cognitive Sciences, Institute for Research in Fundamental Sciences (IPM), Tehran, Iran.

## Abstract

Visual attention brings significant progress for Convolution Neural Networks (CNNs) in various applications. In this paper, object-based attention in human visual cortex inspires us to introduce a mechanism for modification of activations in feature maps of CNNs. In this mechanism, the activations of object locations are excited in feature maps. This mechanism is specifically inspired by attention-based gain modulation in object-based attention in brain. It facilitates figure-ground segregation in the visual cortex.

Similar to brain, we use the idea to address two challenges in salient object detection: gathering object interior parts while segregation from background with concise boundaries. We implement the object-based attention in the U-net model using different architectures in the encoder parts, including AlexNet, VGG, and ResNet. The proposed method was examined on three benchmark datasets: HKU-IS, MSRB, and PASCAL-S. Experimental results showed that our inspired method could significantly improve the results in terms of mean absolute error and F-measure. The results also showed that our proposed method better captured not only the boundary but also the object interior. Thus, it can tackle the mentioned challenges.

## 1. Introduction

Saliency detection refers to detection of most prominent objects on a visual scene. This is inspired by the ability of human eyes to distinguish salient objects from background. Human visual saliency detection is established by two mechanisms: bottom-up saliency and top-down attention [1]. The bottom-up saliency is stimulus-driven attention and formed based on the low-level features, including color, intensity, orientation, texture. It is computed in the feedforward hierarchy of visual cortex. The bottom-up processing initializes and contributes toward visual perception during feedforward computations until an early representation of the object is formed. On the other hand, top-down attention is goal-oriented and refers to the internal guidance of attention based on prior knowledge and the viewer's intention [2, 3].

Motivated by biological vision, research on Salient Object Detection (SOD) and segmentation has been ongoing for the past three decades in computer vision [4, 5]. SOD methods aim to extract areas of salient objects in a visual scene with clear boundaries. Despite the great achievements in

SOD by Convolutional Neural Networks (CNNs) [6, 7], there still exist several challenges. The challenges of SOD include gathering distinct components of an object, ignoring its interior boundaries while segregating its exterior boundaries from cluttered background.

In contrary, Human Visual System (HVS) overcomes these challenges in everyday life. Analysis similarities and differences of CNNs and HVS may motivate us to a solution for overcoming the challenges. The feedforward process in CNNs resembles the overall function of the computational models of the primary HVS [8, 9]. In both systems, different basic features (color, orientation …) are extracted from a visual stimulus, where the features are processed in parallel and separately in different parts. CNNs extract various features by different convolutional filters in different feature maps, and HVS also extracts them in different parts of the visual cortex. However, HVS employs an explicit mechanism to form a unified representation of the whole object, but CNNs do not include such an explicit one. Attention-based feedback constitutes this mechanism in HVS[10-12].

This mechanism is related to object-based attention [13]. Visual attention belongs in certain features, objects, or locations [14], while object-based one corresponds with a high-level abstraction of an object in Brain [15, 16]. When humans attend to some part of an object, representation of the whole object is enhanced in the brain based on top-down feedback. This explicit feedback first helps the model to better segregate objects from background. It also facilitates the perception of an object as a unified whole, regardless of its features and components. It accordingly develops the concept of object-based attention, feature binding, and perceptual grouping in brain [11, 17-19].

Many studies have introduced different object-based visual attention models in computational vision models [13, 20]. Object-based visual attention has also long been suggested for computer vision [21, 22]. During recent years, many works incorporated visual attention into different CNN models [23, 24]; however, a few works explicitly focused on object-based visual attention in CNNs [20, 25, 26]. Spatial and feature-based attention were widely used in CNNs [27], while the object-based one has been neglected. In this paper, we introduce a method to incorporate the object-based attention feedback into the CNNs for SOD. In specific, we implement the neuro-inspired idea in the U-net model [28] in order to examine whether the mechanism could help the model to tackle SOD challenges.

### 1.1 Overview of our inspiration

The Feature Integration Theory (FIT) [11 ,10] and the role of attention in processing and plasticity is the inspiration source of our proposed method. This theory introduced one of the most significant models of human visual attention. It proposes that the attention to an object involved in separate processes. In the first stage of visual perception, different basic features (color, shape …) are extracted from the visual perceived stimulus, where the features are processed in parallel and separately as a pre-attentive stage. This stage occurs early in perceptual processing before we become conscious of the object. The later stage is the focused attention, where individual features of an object are combined to perceive the whole object. This type of attention is called object-based attention. The pre-attentive phase is mainly based on bottom-up processing, while the attention phase relies on the top-down process.

Object-based attention in HVS and its effect on the processing and learning [29] is the main source of our inspiration. We introduce the explicit object-based feedback mechanism that distinguishes the activity of object-viewer neurons from peripheral ones. We manually excite the relative activity of the object-viewer neurons. This excitation is inspired by the attention-based gain modulation of neuron's response [29, 30] in HVS. It helps the higher-layer neurons to better understand and distinguish between the inner and outer regions of the object.

We implement this mechanism in CNNs based on a local feedback module into the mid-layers of the model. Inspired by object-based attention, this module excites the object-viewing neurons to modify their activations to increase the separability of objects and background [31, 32]. The proposed method assumes that object-based attention information is available from the location of objects in a binary map. It is a feasible assumption where saliency and priority maps in HVS [33] include such a binary map. However, the implementation of this assumption is controversial. Despite the works adding an auxiliary network or global feedback in iterative mode, we simply substitute segmentation ground-truth for the object locations in the scene.

## 2. Proposed method

We introduce our neuro-inspired method in U-net to address the challenges in SOD: distinguishing a salient object, including gathering distinct components of an object, ignoring its interior boundaries while segregating its exterior boundaries from cluttered background. We previously suggested the similar approach in general object detection [34], while we adapt it here for segmentation based on some modifications: replacing detection ground-truth with the segmentation one and several minor changes. First, we explain the method from machine learning perspective and then discuss the vision neuroscience motivations behind it.

Our method only works in the learning process. We neither change the U-net architecture nor the segmentation process. We just manually excite the activations of object locations in the feature maps. This auxiliary excitation is applied at the beginning of learning, while gradually reduced with the learning proceeds. It is reduced to zero in the final epochs of learning to adapt the model to work without assisted excitation in the testing phase. The assisted excitation of activations helps the network to distinguish the foreground and salient objects from background. We extract the locations of objects from segmentation ground-truth and excite their corresponding activations in the feature maps.

These excitations guide the model through the segmentation of objects from background. We call our proposed method as Assisted Excitation (AE). However, Ground-truth is only available during training, and thus our final trained model should not count on it. We adapt the model to work without the need for AE, inspiring by curriculum learning [35]. The curriculum learning framework suggests that learning begins with easier tasks, while the more complex ones are gradually considered. It is inspired by human developmental learning where human infant first learns easier tasks and gradually acquires the ability to perform complex tasks. Interestingly, the same applies to neural network learning, curriculum learning suggests. When the problem is non-convex, and the stochastic gradient may fall into a bad local minimum, this learning framework argues that begin with easier tasks and continue with more complex ones. It yields better convergence in terms of the quality of local minima and generalization.

## 2.1. Assisted Excitation using Ground-Truth

The assisted excitation works as a built-in module in CNNs. It manipulates neural activations in the feature maps of a CNN as follows:

(1) $$a^{L+1}_{(c,i,j)} = a^{L}_{(c,i,j)} + \alpha(t)e_{(c,i,j)}$$

where $a^L$ and $a^{L+1}$ are activation tensors at layers $L$ and $L+1$, and $e$ is the excitation matrix broadcasted to all feature maps in a tensor, and $\alpha(t)$ is the curriculum factor works as a function of epoch number t. Also (c; i; j) refer to channel number, row, and column. During training, $\alpha(t)$ begins with a non-zero value while gradually decays to zero. Our proposed excitation, $e$, is a function of activations and ground-truth computed based on the following procedure. We first create a binary segmentation map from ground-truth, as follows:

(2) $$g(i,j) = \begin{cases} 1, & \text{object exists in cell}(i,j) \\ 0, & \text{no object exists in cell}(i,j) \end{cases}$$

Next, the matrix $e(:, i, j)$ get maximum in through all channels (depth of tensor) of $a^l(:, i, j)$. We compute the excitation matrix $e$ as follows:

(3) $$e_{(:,i,j)} = g(i,j)\max(e_{(:,i,j)})$$

Figure 1 illustrates our AE module in more detail. Moreover, Figure 2 presents how we use the AE module in U-net.

## 2.2 AE module in U-net

The AE module can be incorporated on each tensor map of a CNN. We used this module in the encoder and concatenation parts of the U-net model during training. The schematic overview of the proposed model is depicted in Figure 2.

We set the curriculum factor $\alpha = 0$ as if the AE module is removed in the inference phase. Since we gradually decrease the factor during the last epochs of training, our model learns to work without the help of segmentation ground-truth.

In practice, our model architecture is identical to U-net during inference. Our trained model differs from the standard U-net model only in the trained weights. Moreover, the inference time remains identical to the U-net model; however, we achieve better accuracy.

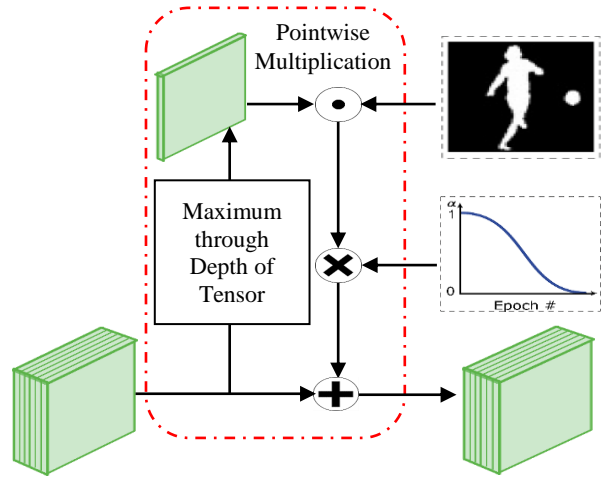

Figure 1 – Illustration of the AE module. An activation tensor is input to this module. First, the maximum matrix is calculated. It then masks the excitation matrix based on the binary segmentation ground truth. The excitation matrix is computed by multiplying the masked matrix and curriculum factor. The excitation matrix is finally broadcasted and added through each feature map in the tensor and passed on the next stage.

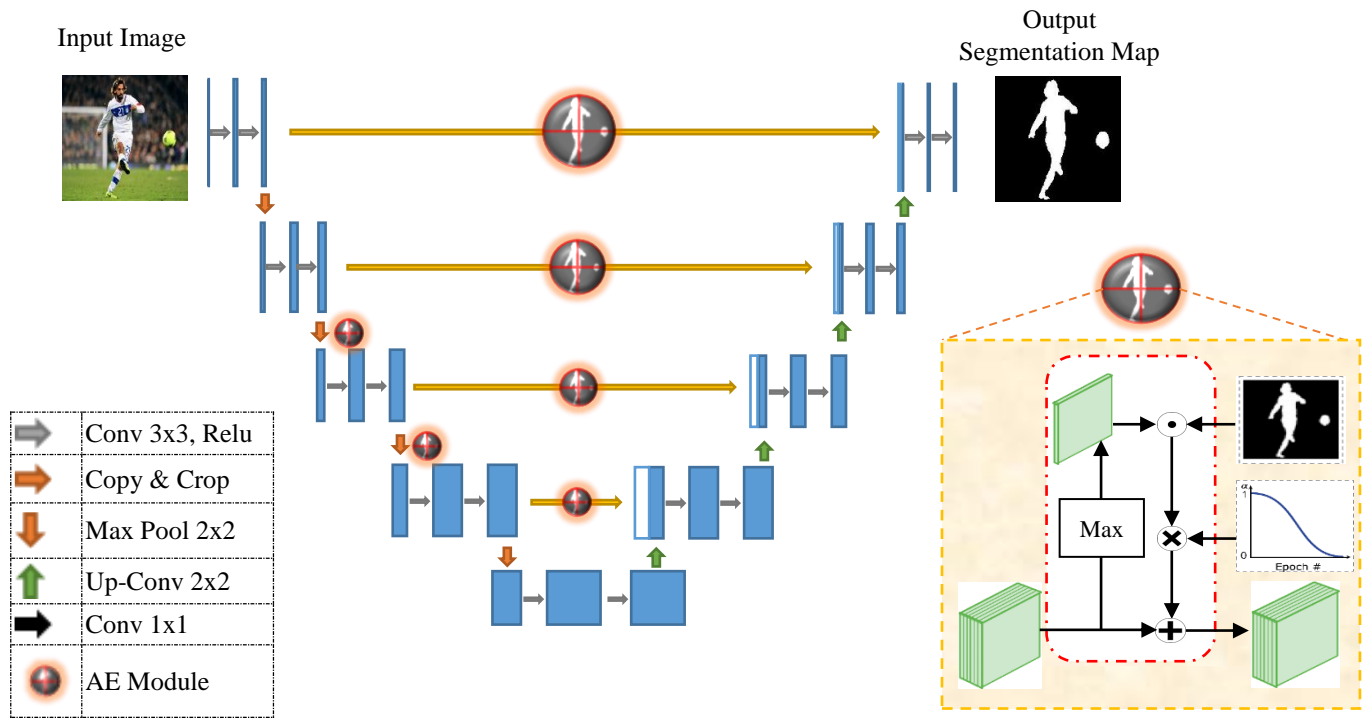

Figure 2. Attention-based AE module in U-net architecture. This module is incorporated between some encoder stages and some concatenation stages.

## 3. Attention-based Assisted Excitation, from a vision neuroscience perspective

In this section, we describe the similarities and differences of the proposed method with the mechanisms involved in the visual cortex of the brain and visual attention. Our proposed method does not model the visual attention system but inspired by it. We first provide an overview of the main similarities of the proposed method with its inspiration. We also get a closer look at these similarities and analyze the correspondence between the components of the AE module and the mechanisms in the visual attention. According to this functional similarity, we complete the name of our proposed method as Attention-based Assisted Excitation (AAE). Finally, we explain the main differences between the real mechanism in brain and ours.

### 3.1 General Similarities

The feedforward process in CNN is similar to the overall performance of the primary visual cortex [36, 37]. In both systems, edge and contour detection, contour grouping, and object boundary formation occur in bottom-up computation. Receptive fields of visual neurons get gradually larger and gradually merged through the hierarchy to include the whole object within. Accordingly, a neuron in the highest layer can view the whole object, classify, and identify it. CNNs work this way; however, visual cortex exerts more diverse mechanisms to perceive objects [11, 38]. One of these mechanisms includes object-based attention formed by top-down feedback in HVS [15].

Our proposed AAE introduces the object-based attention mechanism into the training of a CNN. In the learning phase, a local feedback module is incorporated into the mid-layers of the model. Inspired by object-based attention, this module excites the object-viewing neurons to modify their activations to increase the separability of objects and background. When we manipulate the activations of whole object regions, similar to the process involved in object-based attention, it helps the CNN to group different features and parts of an object into a unified whole. Beyond object-based attention, this integrity mechanism also called feature binding and perceptual grouping [11, 39] in neuroscience. The Gestalt's view [11, 18, 39] to these mechanisms implies that objects are perceived not simply as a set of different parts but as a holistic entity in brain.

Putting it all together, the visual cortex employs two auxiliary mechanisms for better figure-ground segregation, including lateral connection and object-based attention through top-down feedbacks [40, 41], while this mechanism is neglected in CNNs. The general mechanism of the proposed method is similar to the interaction of bottom-up and top-down attention in the brain [42]. Top-down feedbacks influence bottom-up flow by enhancing responses induced by attention [43]. Inspired by them, we accordingly incorporate the feedback mechanism in the CNN model that distinguishes the activity of object-viewer neurons from peripheral ones. We manually excite the relative activity of the object-viewer neurons. It helps the higher-layer neurons to better understand and distinguish between the inner and outer regions of the object [31, 40].

- **Also inspired by developmental learning**

Our proposed assisted excitation is implemented only during the learning phase of the CNN model and does not exist in the test phase. This is the fundamental difference between our approach and human vision. We adapt the model to transfer from the training phase to the test using the curriculum learning [35] in developmental learning [44, 45]. The process of human developmental learning is from easy to difficult so that the human infant learns simple tasks as he grows up and gradually becomes able to learn more complex cognitive and behavioral tasks.

In particular, we propose a novel implementation of curriculum learning, different from the literature. In basic curriculum learning, input data are presented to the model in order of simple to difficult. In our proposed approach, however, moving from simple to difficult is simulated not by changing the order but by modifying the amount of auxiliary information. Introducing the auxiliary knowledge to the model facilitate its learning at the beginning, but it is gradually diminished while learning proceeds. We adapt the model to work without it in the test phase. Because the knowledge extracted from ground-truth will not be available during the test phase. This is the fundamental difference between ours and the brain, where visual attention actively and permanently guides the eyes for a better understanding of the visual scene. There is, however, no attention module in our model in the testing phase.

## 3.2 Functional and detailed similarities

From a closer look, we discuss the similarities of the proposed method with neuronal mechanisms at the algorithmic level.

- **Different components of AAE corresponds to the mechanisms involved in visual attention.**

As mentioned in part 2.1, proposed AAE consists of several stages: 1. Getting maximum through the depth of tensor, 2. Pointwise multiplication by segmentation ground-truth, 3. Multiplied by the curriculum factor, and 4. Add to all feature maps. Following, we explain the relations between these stages and the mechanism involved in visual attention.

The **first stage of AAE** gets maximum values through all neurons who see the same part of a visual scene (with the same receptive fields). This function simulates the max-like operation of complex cells in the lateral connection of the primary visual cortex [46-48]. These maximum activities are then propagated through the higher levels in the feedforward pathway. We assume that the information is accumulated in an extra memory (analogous to visual working memory [49, 50]). Each neuron in the memory-like unit receives, aggregates, and normalizes [51] the accumulated information of its corresponding neurons (with the same receptive field).

In the **second stage of AAE**, the bottom-up flow of accumulated visual information is multiplied by the binary map of objects in the visual scene. The saliency map and the multiplication operation constitute this stage. **First**, the binary map of object locations corresponds with saliency maps in the brain [53, 52]. However, different and distributed saliency maps are formed in different brain areas, but not a particular one. Indeed, considerable studies suggest that the accumulated saliency map is formed in the frontal eye field in the prefrontal cortex [54, 55]. Prefrontal cortex as the highest computational level in the brain, accordingly plays a significant role in the formation of the major saliency map. It controls the attention and working memory processing through top-down feedbacks [57, 56]. We substitute the segmentation ground-truth for the major saliency map in the brain.

Similarly, we assume the ground-truth in the highest level of training CNNs acts likes the major saliency map in the prefrontal cortex. The top-down feedback from the prefrontal cortex conveys the object locations into earlier layers, which then used for modulation. **Second**, the multiplication operator [43, 58-60] is selected to mimic the modulation effect. This modulation superimposes the object map onto the accumulated visual map (calculated on the feedforward and bottom-up pathway). This way, top-down feedbacks influence bottom-up flows by the response modulation induced by attention.

The **third stage of AAE** multiplies the curriculum factor by the resultant refined visual map. The curriculum factor is inspired by the process of familiarity or novelty detection [61] in the brain. The process of novelty detection works based on the theory of unexpected reward. The brain is surprised when see a novel object. The neuronal activations signal surprise and novelty, and their surprise-related responses link with errors in estimates of state values, called Reward Prediction Errors (RPEs). RPEs play a crucial role in learning through neuromodulatory system, especially dopamine system [45, 62]. The dopamine system engages in novelty detection mechanism and thus get more memory, attention, and learning involved [64, 63].

As the learning proceeds, the CNN model gets more familiar with the objects in the training set; thus, their novelty gradually decreases and grabs less attention. We model the novelty detection system in the brain through the curriculum factor. We gradually decrease the factor along the way; thus, the effect of the surprising, novelty, and attention signal is reduced through training. We further explain more why and how this signal affects the learning in the brain and our CNN model.

In the **fourth and last stage of AAE**, the resultant visual map refined by object map and curriculum factor is finally added to each feature map of the input channel. The resultant visual map contains non-zero values in the object locations and zero values in the background. Adding it to the feature maps leads to increasing the activations of the corresponding object locations. In the neuroscience viewpoint, the activity of the object-viewer neurons in the channel is enhanced. It can be interpreted as the additive gain modulation [30, 65] induced by object-based attention. Accordingly, the fourth stage superimposes the top-down modulation effect through feedback on the feedforward visual flow. This effect enhances the correlation and synchrony of object-viewer neurons [66], leading to their more activation.

To sum up, we summarize the overall function of the AAE module this way. The neurons whose objects are in their receptive fields are synchronized together [67, 68], and their information in the corresponding working memory of this layer [69] is accumulated [70]. This process resembles the visual working memory [40, 41, 70]. The memory is the place where lateral connection and top-down attention feedback interact with each other. This feedback specifically modulates the object-viewer neurons and increase their amplitude responses [71]. We illustrate these explanations in Figure 3.

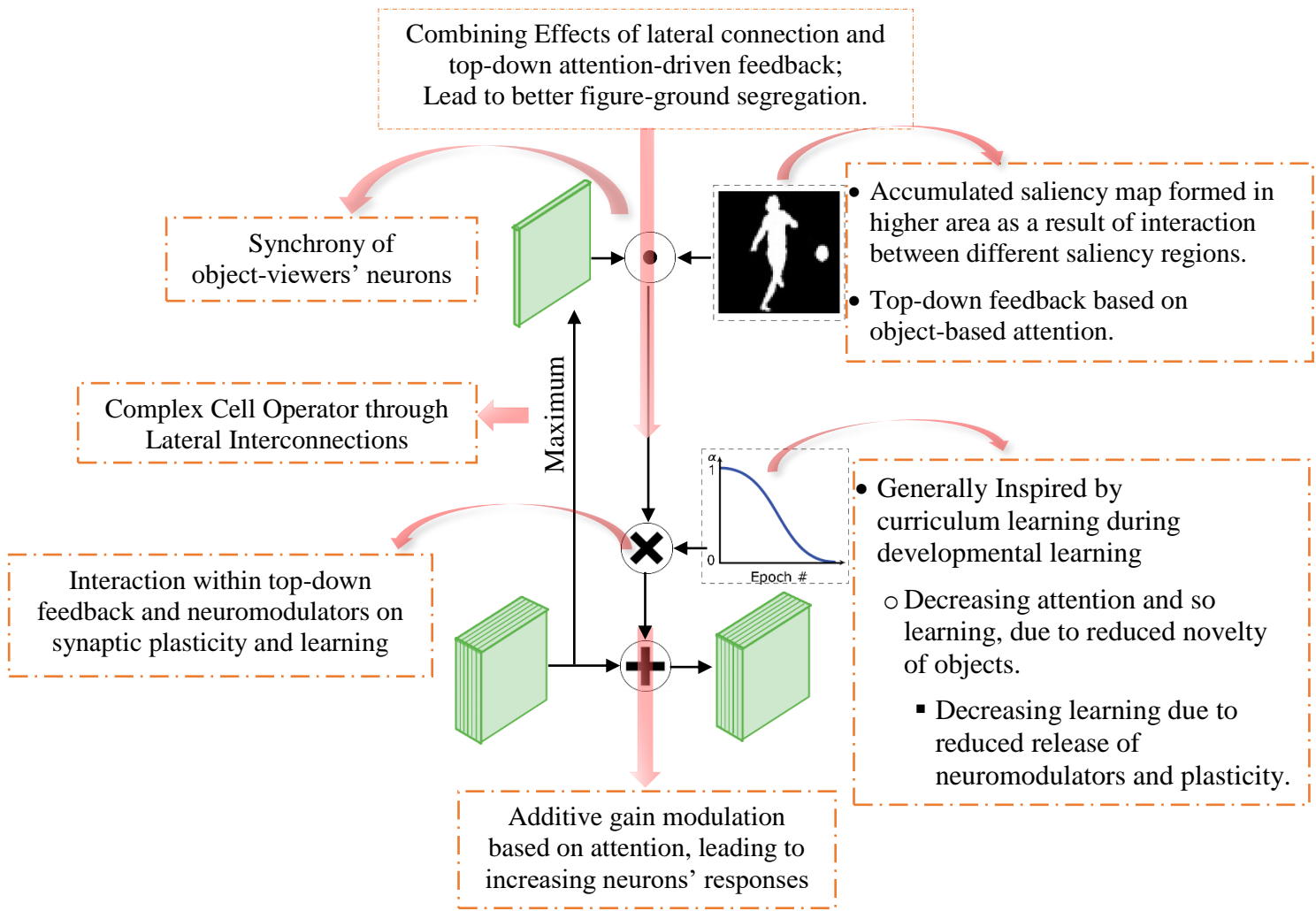

Figure 3- The similarity of our proposed method AAE and the mechanism involved in the visual cortex and visual attention. The black upward arrows mostly correspond to the feedforward processing, and the black downward arrows mostly to the top-down feedback. The horizontal black arrows mostly show the interaction between feedforward and feedback pathways and the modulatory effect of latter.

- **How the proposed AAE improves the training of CNN during backpropagation?**

The role of object-based attention in processing and plasticity (weight changes) is the inspiration source of our proposed method. We include the role of attention in the training phase of CNNs. We excite the object-viewing neurons to enhance their activations to increase the separability of objects and background. However, our AAE module is removed from the model in the test phase due to the lack of ground-truth. A question arises here is how its effect remains and is expected to improve the performance of the model in the test phase.

The more activity of neurons may yield to more plasticity in neuroscience, while it also happens during BackPropagation (BP) [72]. The credit assignment in BP assigns more proportion of weight changes to the more engaged neurons in the output (and the error). The introduced AAE guides the model to train more its weights based on object patterns compared to background. Thus, the positive effect of our proposed attention mechanism is encoded in the weights, embedded in the CNN model and remains in the test phase.

- **Two fundamental factors in plasticity control in AAE**

In our proposed AAE, the curriculum factor affects learning globally on the whole feature map. In contrary, feedback from the object map in ground-truth acts selectively on the object locations. These effects agree with neuromodulation and feedback in the brain. Neuromodulators and feedback are two fundamental factors in learning, where both involved, most learning occurs [29]. However, neuromodulators and especially dopamine are released globally in the brain, but feedback acts selectively. Therefore, based on synaptic tagging, and bi-lateral connection, the corresponding synapses are selected and stimulated to excite synaptic plasticity [73 ,29].

Extending the role of attention to the training phase helps to use attention-based selection and enhancement of neuronal responses [74]. Moreover, attention can also influence plasticity in multiple ways. Attention and neuromodulators are closely intertwined [75]. Attention improves the plasticity through neuromodulators and also by response enhancement that causes more Hebbian learning. When a subject observes a novel object, it grabs more attention, and in addition to novelty prediction by dopamine, more neuromodulatory signals, e.g., Norepinephrine or Acetylcholine release [76]. These signals besides the enhanced response by attention cause more plasticity and synaptic weight changes. As an object becomes more familiar, this novelty signal decreases, and plasticity decreases. It should be noted that though in infants, puberty plasticity is more emphasized, but it continues in adulthood, and there is evidence for plasticity in visual cortex in adulthood [45].

### 3.3 Fundamental Differences

There is a fundamental difference between real attention mechanisms and ours. Our AAE module works during the training phase of CNN while we removed it in practice. However, attention is actively involved in the visual cortex. Another difference is related to the supervised binary segmentation map in our proposed method. We extract the object locations from segmentation ground-truth while this supervised knowledge is not available in the brain. Instead, several saliency maps are distributed and computed in different regions across the cortex and in specific visual cortex. These saliency maps guide our attention towards salient objects in the visual field.

## 4. Experimental Results

In this section, we conduct experiments to evaluate the performance of our proposed AAE+U_Net on the salient object segmentation benchmarks in terms of F-measure and MAE.

### Datasets

Three widely used saliency detection datasets are used to demonstrate the effectiveness of the proposed method, including HKU-IS [77], MSRA-A [78], MSRA-B [79], and PASCAL-S [80].

- HKU-IS [77] dataset has 4,447 images with complex scenes. Most of them contain multiple disconnected salient objects with relatively diverse foreground-background appearance.
- MSRA-A [78] dataset has 20,840 images. Each image has a clear, unambiguous object. MSRA-B is a subset of MSRA-A and has 5000 images. Compared with MSRA-A, MSRA-B [79] is selected from the consistent bounding box labels of the original dataset. Most of the images have only one foreground salient object and clear background, thus has less ambiguity.
- PASCAL-S [80] dataset has 850 natural images selected from the validation set of PASCAL VOC2010 segmentation challenge. It contains cluttered backgrounds and multiple foreground objects and thus more difficult for the salient detection task than the others.

### Implementation Details

We used the different implementations of U-net based on [81]. Different architectures were used in the encoder parts of U-net, including Alexnet, VGG, ResNet [83, 82].

### Evaluation Metrics

There are several ways to measure the agreement between model predictions and human annotations. We evaluate the quantitative performance based on these measures: Precision-Recall, F-measure, and Mean Absolute Error (MAE). These measures are calculated as follows:

$$\text{Precision} = \frac{TP}{TP+FP} \quad , \quad \text{Recall} = \frac{TP}{TP+FN}$$

, where *TP, TN, FP,* and *FN* denote true positive, true negative, false positive, and false negative.

Precision-Recall is calculated based on the binarized saliency mask and ground-truth. F-measure [84] considers both Precision and Recall by computing the weighted harmonic mean:

$$F_\beta = \frac{(1+\beta^2)\text{Precision} \times \text{Recall}}{\beta^2 \text{ Precision} + \text{Recall}}$$

, where $\beta^2$ is empirically set to 0.3 to emphasize more on precision. An adaptive threshold is first used to segment the saliency map $S$ and obtain the Precision-Recall, and the F-measure score is finally calculated. Although this measure does not explicitly consider true negative pixels. To address this problem, MAE is also commonly used. MAE is calculated based on the average pixel-wise absolute error between the normalized map $S$ and the ground-truth mask $G$.

$$\text{MAE} = \frac{1}{W \times H} \sum_{i=1}^{W} \sum_{j=1}^{H} |G(i,j) - S(i,j)|$$

,where $W$ and $H$ denote width and height of the saliency map, respectively. MAE is complementary to F-measure, which calculated the absolute distance between the results and the ground truth.

**Results**

We evaluated the performance of our proposed method attention-based AE in U-net model in MSRB, HKU-IS, and PASCAL-S datasets. The results are reported and compared in Table 1. Moreover, the qualitative results are also compared in Figure 4.

Table 1. The performance of AAE-based U-net with the different encoder architectures are compared to original U-net in MSRB, HKU-IS, and PASCAL-S datasets. The results are also compared to several state-of-the-arts results.

| Method | | MSRB | | HKU-IS | | PASCAL-S | |
|---|---|---|---|---|---|---|---|
| Model | Encoder | MAE | *F*-measure | MAE | *F*-measure | MAE | *F*-measure |
| U-Net | AlexNet | 0.144 | 0.813 | 0.081 | 0.804 | 0.149 | 0.719 |
| | VGG-16 | 0.107 | 0.859 | 0.059 | 0.849 | 0.126 | 0.762 |
| | ResNet | 0.112 | 0.850 | 0.060 | 0.846 | 0.123 | 0.767 |
| AAE-based U-Net | AlexNet | 0.132 | 0.822 | 0.077 | 0.811 | 0.140 | 0.730 |
| | VGG-16 | 0.095 | 0.865 | 0.053 | 0.856 | 0.119 | 0.775 |
| | ResNet | 0.111 | 0.853 | 0.057 | 0.850 | 0.121 | 0.771 |
| NLDF [85] | VGG-16 | 0.048 | 0.911 | 0.048 | 0.902 | 0.099 | 0.826 |
| EARN [86] | VGG-16 | 0.036 | 0.933 | 0.046 | 0.912 | 0.095 | 0.845 |

The experimental results in Table 1 showed that using the deeper and wider encoders, i.e., the VGG and ResNet in the U-net model performed better than the Alexnet encoder in all datasets for all criterion. The comparison of the VGG and Resnet-based encoders in the U-net model showed that neither was significantly superior to another.

The experimental results confirmed that our proposed AAE could improve the performance of U-net with different encoders. In all benchmark datasets, the F-measure values of AAE-based U-net were increased compared to the original U-net. The MAE values were also reduced by our proposed method. However, the proposed Attention-based AE module could not achieve state-of-the-art compared to [85, 86]. In particular, our proposed AAE module could improve the performances of all U-net architectures. The AAE in the Alexnet-based U-net improved the basic model significantly. The AAE in the VGG and Resnet-based U-net also improved the performances compared to the basic models. However, these improvements are not significant.

The qualitative results of SOD in Figure 4 generally showed that AAE could increase both numbers of TP and FP values. The FN values are also reduced in most cases. The quantitative results of SOD in Table 1 showed that the increasing rates of TP values were more than FP values. The FN values were also reduced. Therefore, both the precision and recall values were improved, and accordingly, F-measure values were improved.

We also reported the negative results in the last row of Figure 4. None of the U-net models was able to detect the salient object in the image of basketball hoop; instead, the surrounding trees were highlighted as the salient objects. Since the image of basketball hoop is extremely rare in HKU-IS dataset, the model could not learn and detect the saliency of this object. Moreover, the saliency datasets were tagged by humans, where rareness and cognitive experiences influence human's tagging. However, the basic CNN models, i.e., U-net could not learn and capture this high-level abstraction from the limited dataset. Accordingly, there still exists a significant difference in the saliency perception between humans and CNNs. However, combining high-level heuristics such as rareness and also big enough image dataset for training CNNs may reduce such negative results.

| Input Image | U-net (AlexNet) | U-net (AlexNet) + AAE | U-net (VGG) + AAE | U-net (ResNet) + AAE | Ground Truth |
|---|---|---|---|---|---|
| 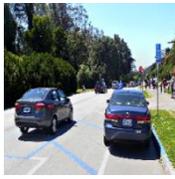 | 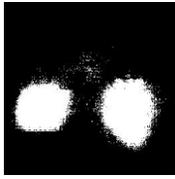 | 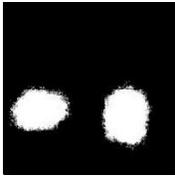 | 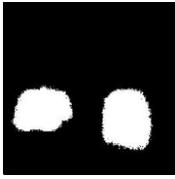 | 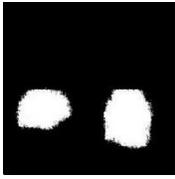 | 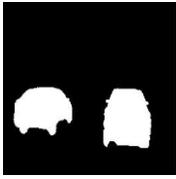 |
| 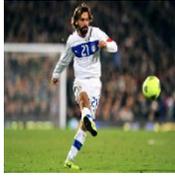 | 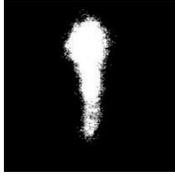 | 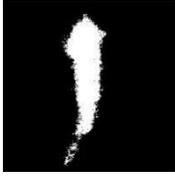 | 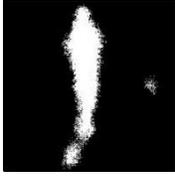 | 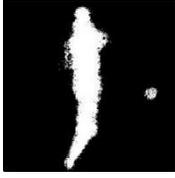 | 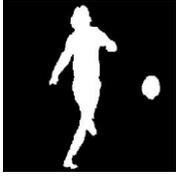 |
| 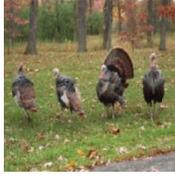 | 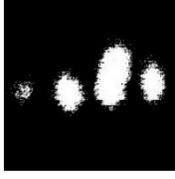 | 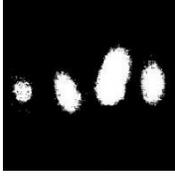 | 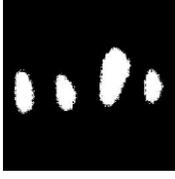 | 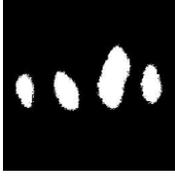 | 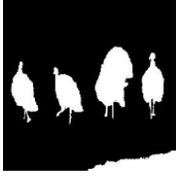 |
| 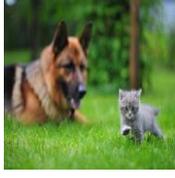 | 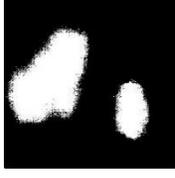 | 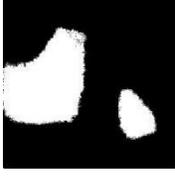 | 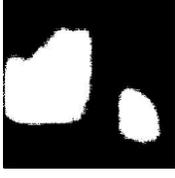 | 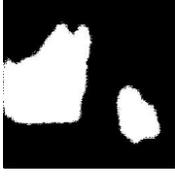 | 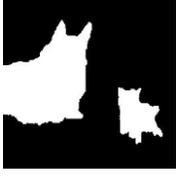 |
| 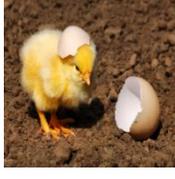 | 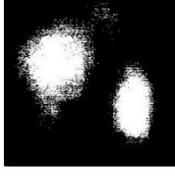 | 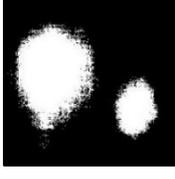 | 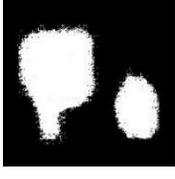 | 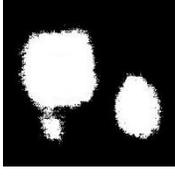 | 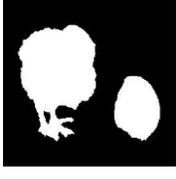 |
| 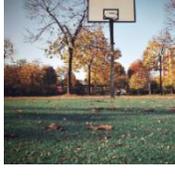 | 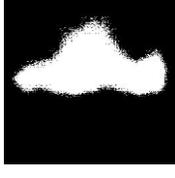 | 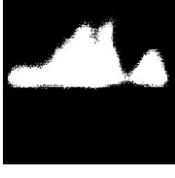 | 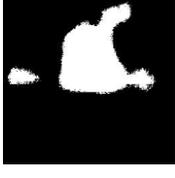 | 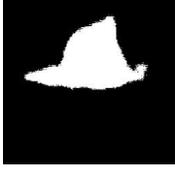 | 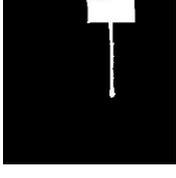 |

Figure 4 - The qualitative results of our proposed attention-based AE is the U-net with different encoders are visually compared.

## Conclusion

Object-based visual attention motivated us to introduce the similar mechanism in CNNs and in particular in the U-net model for SOD. We specifically are inspired by the attention gain modulation in the visual cortex and implemented it as the AAE module. It can be incorporated into a wide variety of CNNs. This mechanism implies that the foreground objects are perceived as a coherent region distinct from the background.

However, the difference between the real brain mechanism and ours is in two folds: 1. Our AAE module works during the training phase of U-net, while we remove it in practice. However, attention is actively involved in the visual cortex. 2. We extract the object locations from segmentation ground-truth while this supervised knowledge is not available in the brain. Instead, several saliency maps are distributed and computed in different regions across the cortex and in specific visual cortex. These saliency maps guide our attention towards salient objects in the visual field.

From a machine learning viewpoint, the proposed idea is a complement for end-to-end learning based on error backpropagation on the problems related to object localization. In fact, one of our innovations is the use of ground-truth not only as the target of error backpropagation but also as a method for direct modification of the intermediate representations in CNNs. The direct manipulation of activations in the feature maps of the CNN leads to better representational learning.

## Acknowledgment

The authors would like to acknowledge the contribution of Mohammadmahdi Derakhshani, Amir Hossein Shaker, Omid Mersa, and Mohammad Amin Sadeghi in the previously published conference paper "*Assisted Excitation of Activations: A Learning Technique to Improve Object Detectors*" [34].